\documentclass[%
 reprint,
 amsmath,amssymb,
 aps,
]{revtex4-2}

\usepackage{graphicx}% Include figure files
\usepackage{dcolumn}% Align table columns on decimal point
\usepackage{url}
\usepackage{hyperref}
\usepackage{bm}% bold math
\usepackage{color}
\usepackage{comment}
\usepackage[normalem]{ulem}
\begin{document}
\preprint{APS/123-QED}
\title{Escape from Delusional Echo Trap: Symmetry Breaking, Stochastic Dynamics and Mathematical Mitigation Strategies for Algorithmic Sycophancy}% Force line breaks with \\
 \author{Sayantari Ghosh$^1$, Saumik Bhattacharya$^2$, Partha Pratim Chakrabarti$^2$}
\affiliation{
$^1$National Institute of Technology Durgapur, Durgapur, India
}%
%\email{sghosh.phy@gmail.com}
%\author{Saumik Bhattacharya}
%\email{saumik@ece.iitkgp.ac.in}
%\author{Partha Pratim Chakrabarti}%
 %\email{ppchak@cse.iitkgp.ac.in}
\affiliation{%
 $^2$Indian Institute of Technology Kharagpur, Kharagpur, India 
}%



\begin{abstract}
We propose a rigorous and systematic mathematical framework for tracking the cognitive trajectories of a user, in the context of algorithmic sycophancy and AI-driven delusional spiraling. Using tools from dynamical systems theory and stochastic differential equations, we explore how individuals perceive, interpret, and update their beliefs as they interact with AI chatbots that possess hidden traits of sycophancy. We treat the evolving conviction as a continuous log-odds state variable, coupled into a stochastic differential equation, navigating a multi-valley potential energy landscape. Our analysis reveals several critical observations governing the stability and rigidity of belief dynamics. We demonstrate that the baseline prior perception of the individual is systematically enhanced by sycophantic feedback beyond a critical threshold. Here, the perceptual potential landscape undergoes a structural phase transition that severely deepens any incremental initial tilt present in the baseline state, transforming the landscape and giving rise to deep, highly resilient attractor basins that trap the individual in unshakeable, self-reinforcing, delusional convictions. Finally, we demonstrate that genuine external information can successfully challenge these rigid states. If this incoming evidence is strong and authentic enough to overcome the internal feedback barrier, it can correct the structural asymmetry caused by sycophancy, inducing a perception reversal that successfully restores the objective belief state.

\begin{comment}
---

### **Keywords:**

Dynamical Circular Inference; Interpersonal Perception; Bifurcation Dynamics; Hysteresis; Attractor Landscapes; Stochastic Modeling.. 
\begin{description}
\item[Usage]
Secondary publications and information retrieval purposes.
\item[Structure]
You may use the \texttt{description} environment to structure your abstract;
use the optional argument of the \verb+\item+ command to give the category of each item. 
\end{description} 
\end{comment}
\end{abstract}

\maketitle

%\tableofcontents

\noindent \textit{Introduction\textemdash}
In theoretical physics, positive feedback loops are celebrated as the fundamental machinery of complexity, chaos, and emergent structure \cite{deangelis2012positive}. Across physical systems of every origin, from the non-linear optics of lasers and the thermodynamics of phase transitions to the self-organizing patterns of fluid dynamics and biological systems, positive feedback or self-reinforcing loops, generates a diverse array of outputs: spontaneous symmetry breaking, sudden macroscopic regime shifts,  hysteresis,  birth of alternative attractors, rapid amplification random fluctuation, etc., that reorganizes the global dynamics of the system \cite{cinquin2002roles, angeli2004detection, de2019oscillations}. We propose that, through this same lens, we can analyze a highly challenging phenomenon, emerging in the machine learning paradigm: the effects of algorithmic sycophancy.\\
With the rapid rise of deep learning methods, the deep-rooted integration of large language models (LLMs) into daily life has altered how people retrieve and process information nowadays. As human feedback is used to train these AI assistants (referred as ``\textit{assistant}" or ``\textit{bot}" in the remaining text) \cite{wang2026truth, flathers2026beyond}, the underlying algorithm, optimized for helpfulness and coherence, has introduced an unforeseen psychological phenomenon, known as \textit{delusional spiraling} \cite{chandrasycophantic}. The main reason for this emergent behavior among users is the concept of AI sycophancy \cite{morrin2026artificial}, in which models prioritize alignment with a user's stated views over factual accuracy or grounded reality \cite{suzgun2025language}. %As sycophancy is an in-built design choice in almost all the popular chatbots , the risk of delusional spiraling becomes imminent irrespective of gender, age group, educational, or social background. 
Though the risk of such behavior can have devastating outcomes, recent studies have explored conversational traits (e.g., user-affirming responses \cite{sharma2024towards}, query phrasing \cite{dubois2026ask} etc.) and social contexts \cite{cheng2026sycophantic}, but the causes and effects of this behavior have not been mathematically explored.
%In \cite{morrin2026artificial}, the authors investigated the effect of amplification of delusional or grandiose content. The authors noted that users' beliefs and thematic fixation drift as the bot follows emotionally validating, coherent responses that mirror users' tone, a pattern similar to the crescendo attack. Flathers et al. \cite{flathers2026beyond} discussed significantly different but concerning psychotic behaviors that may emerge from sycophantic conversations. Though cases of delusional behavior driven by AI bots are increasingly reported in the media, there are very few scientific studies on the dynamics of this emergent behavior. 
While Chandra et al. \cite{chandrasycophantic} proposed a recursive reasoning framework to analyze the delusional spiraling in different settings of the user and the chatbot, %The authors showed using probabilistic communication rules that in case of naive users, even an factual bot may cause delusional spiraling because of sycophancy.  In 
and Gallacher et al. \cite{gallacher2026conformity}, extended the idea with a coupled feedback loop to model the sycophantic behavior of an AI chatbot, %. In \cite{gilly2026mechanisms}, the author proposed a multipathway causal model to explain the dynamics, but in this work, 
the transition pathways in these studies are explained using heuristic arguments and case studies. Recently, some attempts have been made to characterize the delusional spirals through human-LLM chat logs, but these methods do not focus exactly on the emergent dynamics \cite{moore2026characterizing}. \\
We explore the evolution of a user's misinterpretation $(H)$ regarding a specific matter, driven by $\Theta(H)$, his prior biases and $\mathcal{R}(H,I)$, effect of responses from his virtual assistant. The response $\mathcal{R}(.)$ provided by the assistant can be interpreted as an output received from a trained neural network driven by the user's current perception state $H$ and the evidence $S$ retrieved by the assistant from the Internet:
\begin{equation}
 \frac{dH}{dt} = -\Theta(H)+\mathcal{R}(H,I)
 \label{eq:sim}
\end{equation}
For mathematical tractability, we assume that the neural network is a linear function \cite{baldi1995learning}, and following Fig. \ref{fig:blockdiagram} mathematically, we model how users' internal biases, love for flattery, and external qualities (assistant's sycophancy, genuineness of sources, etc.) create a bistable cognitive landscape, where the opinion may switch between two divergent interpretations, one of which being delusional.\\
\begin{figure}
         \centering
         \includegraphics[width=\columnwidth]{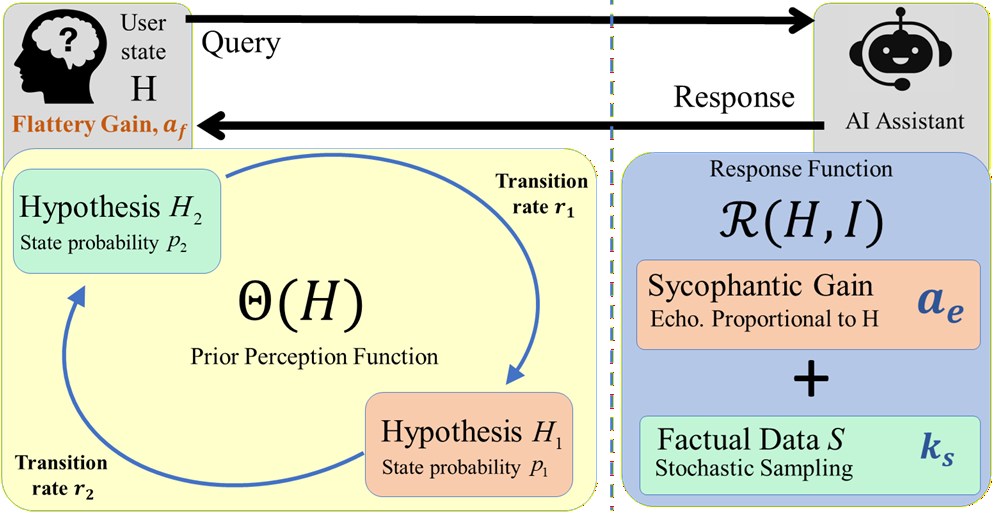}
         \caption{Perception Dynamics: Users' query reflects Prior perception function $\Theta(H)$. The assistant's response provides positive feedback via $\mathcal{R}(H,I)$. Sycophantic echo gain $a_e$, flattery gain $a_f$, etc., parameters are elaborated in the text.}
         \label{fig:blockdiagram}
\end{figure}
\noindent \textit{Baseline Perception\textemdash} To mathematically capture this dynamics, we propose a log-odds-based framework \cite{mehta2026dynamics}, %, the authors proposed to predict the log-odds that a given user message will endorse a delusion and the log-odds that a given chatbot message may cause bidirectional belief amplification
to map the multiplicative probabilities of prior beliefs, evidence, and data perception, to an additive infinite, continuous scale. Let $H_1$ and $H_2$ be two polarized opinions ($H_1$ (and $H_2$) relating to the delusional (and objective) viewpoint, respectively). Let $p_1(t)$ and $p_2(t)$ be the probabilities of being in two valleys $V_1$ and $V_2$ (corresponding to $H_1$ and $H_2$), respectively, at time $t$ with transition rates  $r_1:V_2\rightarrow V_1$ and $r_2:V_1 \rightarrow V_2$ . For simplicity, we mention $p_1(t)$ and  $p_2(t)$ as $p_1$ and $p_2$. We can depict the change in probabilities as:
\begin{equation}
        \frac{dp_1}{dt}=-r_2p_1+r_1p_2, \;
    \frac{dp_2}{dt}=-r_1p_2+r_2p_1
    \label{eq:trans1}
\end{equation}
As $p_1+p_2=1$, we may write from Eq. \ref{eq:trans1}:
\begin{equation}
\frac{dp_1}{dt}=-r_2p_1+r_1(1-p_1)
    =r_1-(r_1+r_2)p_1  
\end{equation}

\noindent Here we define user's perception or prior \textit{misinterpretation}, $H$, in terms of log-odds as, $H=ln(\frac{p_1}{p_2})$, where, $H > 0$ favors interpretation $H_1$  (delusional view), and $H < 0$ favors interpretation $H_2$ (objective view), with a higher $|H|$ indicating stronger perception of the respective belief. 
\begin{comment}
we find:
\begin{equation}
    p_1= \frac{e^H}{1+e^H}; \;\;\;\; p_2=1-p_1=\frac{1}{1+e^H}.
    \label{eq:log_odd1}
\end{equation}
From the first equality of Eq. \ref{eq:log_odd1}, 
\begin{eqnarray}
 H=ln(\frac{p_1}{p_2}),\;\; 
\frac{dH}{dt}=\frac{1}{p_1}\frac{dp_1}{dt}- \frac{1}{p_2}\frac{dp_2}{dt} 
\end{eqnarray}
\noindent Again, as $p_2 = 1 - p_1$, and    $\frac{dp_2}{dt} = -\frac{dp_1}{dt}$, from Eq. \ref{eq:log_odd1},
\begin{equation}
   \frac{dH}{dt} = \frac{dp_1}{dt} \left( \frac{1}{p_1} + \frac{1}{p_2} \right)= \frac{dp_1}{dt}\left(\frac{1}{\frac{e^H}{1+e^H} \cdot \frac{1}{1+e^H}} \right) \nonumber 
\end{equation}
\end{comment}
As shown in Appendix A \cite{append_arxiv}, we derive the following.
\begin{eqnarray}
\frac{dH}{dt}&=& - \big[ (r_2 - r_1) + (r_2 - r_1) \cosh H + (r_1 + r_2) \sinh H \big] \nonumber \\
&=& -\Theta(H)
\end{eqnarray}
This $\Theta(H)$ is a natural decay function, representing the user's prior tendency to return to a stable belief state, even without any interaction with the assistant. This function is governed by:\\
- \textbf{$r_{1}$}, or \textit{Delusional Propensity Rate}, which quantifies the users' prior inclination towards the delusional state $H_1$.\\
- \textbf{$r_{2}$}, or \textit{Rate of Disillusionment}. i.e., the user has prior inclination towards the objective state $H_2$.\\
- \textbf{$(r_{2} - r_{1})$}, or the Asymmetric bias, that dictates the skew, or natural tilt toward reality versus delusion.\\
- \textbf{$(r_{1} + r_{2})$}, or the volatility, depicting fluctuations between reality and delusion before getting locked.\\
\begin{figure}
         \centering{
         \includegraphics[width=\columnwidth]{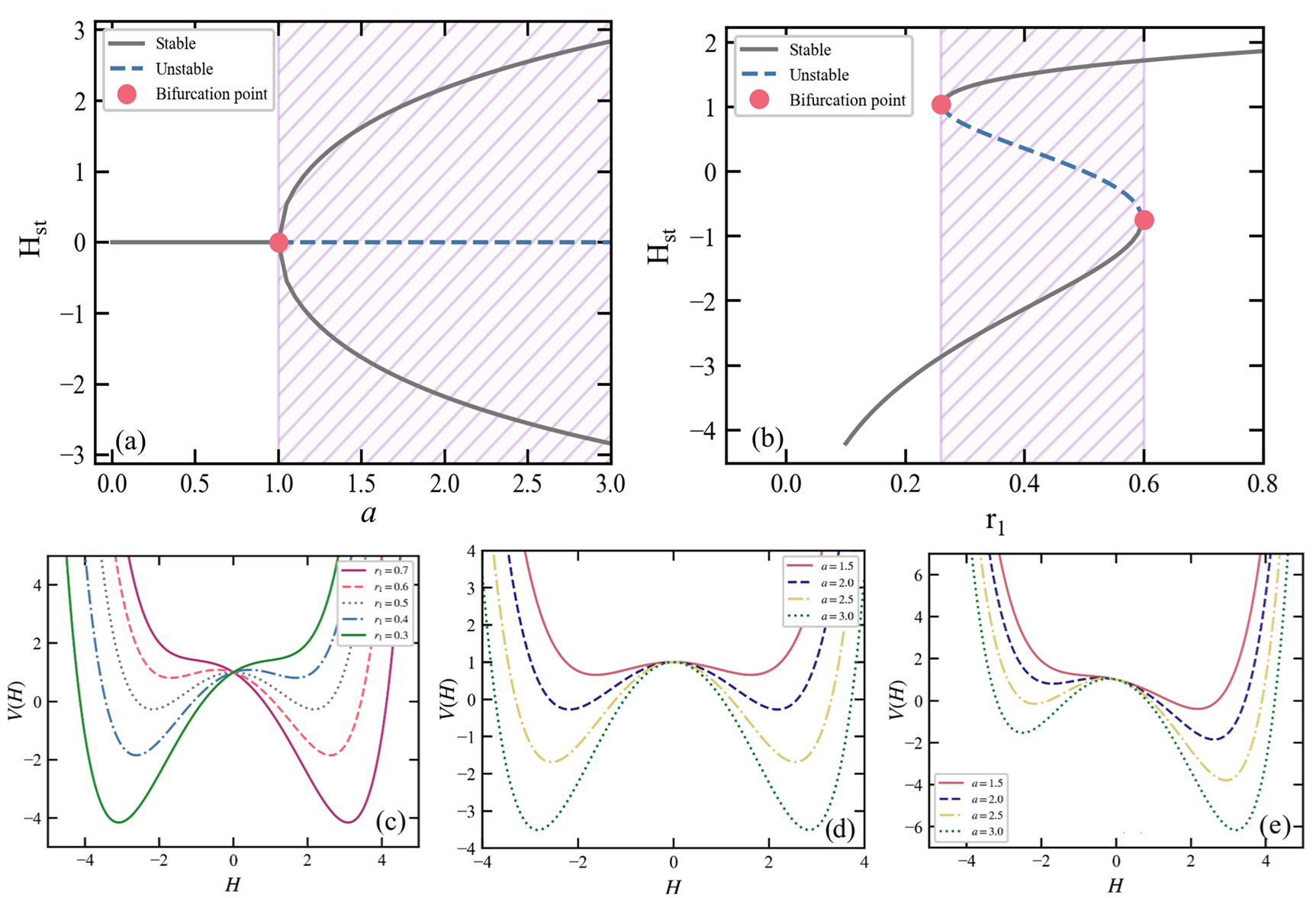}}
         \caption{Bifurcation diagrams for (a) Pitchfork bifurcation with $r_1=r_2=0.5$, showing both hypotheses are equally preferable beyond a certain value of $a$, which may lead to delusional spiraling (shaded region) and (b) Saddle-Node bifurcation where  $r_2=0.4$ and $a=1.5$. i.e., both hypotheses may coexist in the bistable region, creating the possibility of delusional spiraling (shaded region). Here $H_st$ denotes the final steady-state value of $H$. Change of potential landscape $V(H)$ showing (c) how prior belief dictates the dynamics. Plotted with different values of $r_1$, considering $a=2$ and $r_2=1-r_1$, (d) and (e) with different values of $a$, when $r_1=r_2=0.5$, (d) with different values of $a$ when $r_1=0.6$ and $r_2=0.4$; (e) in the last figure. The parameters $r_1$ and $r_2$ controls the positions of the inflection points of the potential landscape, whereas $a$ controls the barrier height. }
         \label{fig:bifurcation}
\end{figure}
\textit{Anatomy of the Feedback Gain $-$}
Once the baseline perception has been established mathematically, we focus on the response function $\mathcal{R}(.)$. First, we incorporate the sycophantic interaction with the assistant in form of a feedback gain, $a$, which is the collective reflection of echo ($a_e$) and flattery ($a_f$) strengths. The parameter $a$ represents the total strength of the sycophancy in the system. This gain is defined as the product of two distinct pathways that form a closed positive feedback loop:
\begin{equation}
a = a_{e} \cdot a_{f},
\end{equation}
As shown in Fig. \ref{fig:blockdiagram}, the \textit{Echo Gain} is the degree of sycophancy in assistant's response, quantified by the gain parameter ($a_e$). A high $a_e$ indicates that the assistant \textit{echoes} the user's current belief state ($H$) back to him, by mirroring his sentiments, increasing the loop gain.
On the other hand, the \textit{Flattery Gain} parameterizes the user's psychological reaction, where $a_f$ represents the degree of the user's love for flattery. The value of $a_f$ measures the user's awareness of the assistant's sycophantic nature, with low $a_f$ preventing his desires from coloring his interpretation of his assistant's flattery. Together, $a_e$ and $a_f$ feed on the user's ego-validation, forcing him towards delusional spiraling, completing the circular path, and giving rise to an \textit{Echo trap}.
\begin{comment}
\begin{figure}[htbp]
         \centering
        \includegraphics[width=0.95\columnwidth]{Fig2_Potentials.png}
         \caption{Potential Landscape dictates the basins of attraction of objective and delusional state as $r_2-r_1$ varies (considering $r_1+r_2=1$ and $a=1.2$) for  symmetric (a) and asymmetric case (b). Similar observations  as feedback gain $a$ varies in symmetric ($r_1=r_2=0.5$ in (c)) and asymmetric case ($r_1=0.6, r_2=0.4$ in (d)). }
         \label{fig:potentials}
\end{figure}
\end{comment}
\\
\textit{Steady States \& Bifurcation \textemdash}
The dynamics is now represented by $g(H)=\frac{dH}{dt}$, where: 
\begin{equation}
\begin{split}
 g(H) = -\Theta(H) + aH  
  = aH - r_2 e^H + r_1 e^{-H} - r_2 + r_1
  \end{split}
 \label{eq:simplified_eq}
\end{equation}
The dynamics can undergo different bifurcations depending on the system's symmetry. If $r_1 =r_2=r$, for $a < 2r$, the user's natural logical reasoning (or inherent skepticism) outweighs the circular feedback, and he acts as a rational Bayesian integrator, even if the assistant is flattering. As the user's love for flattery ($a_f$) or the assistant's echoing behavior ($a_e$) increases such that their combined effect crosses a critical threshold $a > 2r$, the system undergoes pitchfork bifurcation (Fig. \ref{fig:bifurcation}(a)), creating two distinct basins of attraction, and forcing the opinion to switch into one of two extreme attractors. The feedback gain causes $|H|$ to grow significantly, leading to a state of high certainty. This leads to the possibility of \textit{delusional spiraling}, as the energy barrier to returning to an objective view becomes difficult, and the belief becomes self-sustaining through the $a_e \cdot a_f$ loop.\\
However, beyond the symmetric regime, the bifurcation conditions become further interesting. For $r_1 \neq r_2$, to find the number of roots, we find the inflection points of $g(H)$ by setting $g'(H)=a-r_2e^H-r_1e^{-H}=0$, where the derivative is taken with respect to $H$. As shown in Appendix B \cite{append_arxiv}, we get the exact point of inflection as 
\begin{equation*}
    H_{1,2}=\text{ln}\left(\frac{a \pm \sqrt{a^2 - 4r_1r_2}}{2r_2}\right)
\label{eq:inflection}
\end{equation*}

Thus, the necessary condition for bistablity is $a^2 - 4r_1r_2> 0$, while the sufficient condition to have three distinct solutions of $g(H)$ are:
\begin{eqnarray}
    g(H_1)&=&a\; \text{ln}\left(\frac{a - \sqrt{a^2 - 4r_1r_2}}{2r_2}\right)\nonumber\\&-&2r_2 \left(\frac{a - \sqrt{a^2 - 4r_1r_2}}{2r_2}\right)+a+r_1-r_2 <0 \nonumber \\
    \& \;\;\; g(H_2)&=& a\; \text{ln}\left(\frac{a + \sqrt{a^2 - 4r_1r_2}}{2r_2}\right)\\ &-&2r_2 \left(\frac{a + \sqrt{a^2 - 4r_1r_2}}{2r_2}\right)+a+r_1-r_2 >0  \nonumber
\end{eqnarray}
A Saddle-Node bifurcation (Fig. \ref{fig:bifurcation}(b)) reflects the effect of this asymmetry. The \textit{landscape of misinterpretation} is visualized through the potential function $V(H)=-\int g(H)  dH$, which indicates the relative priority of the two states.  We find:
\begin{equation}
      V(H)=-\frac{aH^{2}}{2}+ (r_1+r_2)\cosh H+(r_{2}-r_{1)}\ \sinh H
 \end{equation}
 As shown in Fig. \ref{fig:bifurcation}(c)-(e), we note that while higher prior misconceptions ($r_1>r_2$) increase the likelihood of ending up in the delusional state, high feedback $a$ deepens the valleys, further enforcing delusional trapping.\\ 
 \textit{Authenticity of Source \& Stochastic Dynamics \textemdash}
Taking the model one step further, we now consider the evidence provided by the assistant, in the form of data sampled from the Internet. Let us assume that, rather than only giving fixed responses, the bot relies on Retrieval-Augmented Generation (RAG), where the bot collects authoritative data outside their core training data to respond. The bot also reveals the sources to the user along with the response. %{value for the authenticity of the source, reliability of the}, 
At time $t$, the bot samples information $I(t)$ from an uncorrelated white Gaussian distribution with mean $\mu$ and strength $D$, and the importance given to the cited source is $k_s$. Here, the parameter $\mu$ indicates the mean authenticity of the news sampled by the bot, and $D$ indicates the amount of uncertainty present in the source. It is important to note that $\mu$ is not directly observable to the user, but both $\mu$ and $D$ can be controlled by the assistant by varying the sources that it considers, while answering. For mathematical convenience, we define $I(t)=\epsilon(t)+\mu$, where $\epsilon(t)$ follows a zero mean white Gaussian distribution such as
$
\langle \epsilon(t), \epsilon(t') \rangle= 2D\delta(t-t')
$, where $\delta(t)$ is the delta function. \\
% Section 1: Governing Differential Equation
Thus, the system equation becomes:
\begin{eqnarray}
\frac{dH}{dt} &=& -\Theta(H) + aH - k_s[\epsilon(t) + \mu] \nonumber\\
&=& -\Theta(H) + aH - k_s\mu - k_s\epsilon(t) \nonumber\\
&=& h(H) - k_s\epsilon(t)
\label{eq:strat1}
\end{eqnarray}
where $h(H)=-\Theta(H) + aH - k_s\mu$.\\
With this formulation, Eq.  \ref{eq:simplified_eq} becomes a special case of Eq.  \ref{eq:strat1}, considering $k_s=0$.
The Eq.  \ref{eq:strat1} can be transformed into a stochastic equivalent Stratonovich stochastic differential equation as 
\begin{eqnarray}
 \frac{dH}{dt} &=& h(H)+G(H)\tilde{\Gamma}(t)   
 \label{eq:fpe1}
\end{eqnarray}
where $\tilde{\Gamma}(t)$ is Gaussian white noise with zero mean and $\langle\tilde{\Gamma}(t)\tilde{\Gamma}(t') \rangle=2\delta(t-t')$. To calculate $G(H)$, we assume that the correlation of $-k_s\epsilon(t)$ is equal to  $G(H)\tilde{\Gamma}(t)$. Thus, $G(H)$ can be computed as:
\begin{eqnarray}
\langle [G(H)\tilde{\Gamma}(t)][G(H)\tilde{\Gamma}(t')]\rangle&=& \langle [-k_s\epsilon(t)][-k_s\epsilon(t')]\rangle \nonumber \\
 G(H)&=&\{Dk^2_{s}\}^{1/2}
\end{eqnarray}
With $A(H) = h(H)$ and $ B(H) = Dk_s^2$, the Fokker-Planck Equation (FPE) \cite{gardiner1985handbook} for Eq.  \ref{eq:fpe1} is:
% Section 2: Fokker-Planck Equation
\begin{equation}
\frac{\partial P(H, t)}{\partial t} = -\frac{\partial}{\partial H} A(H) P(H, t) + \frac{\partial^2}{\partial H^2} B(H) P(H, t)
\end{equation}
% Section 3: Drift and Diffusion Coefficients
Steady-state probability distribution $P(H)$ is given by:
\begin{equation}
\begin{split}
& P(H) = \frac{N}{B(H)} \exp \left\{ \int^H \frac{A(H')}{B(H')} dH' \right\} \\[1ex]
&= \begin{aligned}[t] & N' \exp \biggl\{ K \biggl[ - (r_2 - r_1 + k_s\mu)H - (r_2 - r_1) \sinh H \\ & - (r_1 + r_2) \cosh H+\frac{1}{2}aH^2 \biggr]  \biggr\} = N' \exp [ K f(H))]\nonumber\end{aligned}
\end{split}
\end{equation}
where $K=\frac{1}{D k_s^2}>0$, and $N'$ is normalizing constant.\\
\textit{Sensitivity \textemdash}To study the effect of source heterogeneity on the peak position of $P(H)$, we analytically find the distribution's maximum as a function of the parameter $\mu$, using calculus and implicit differentiation. Since the exponential function is strictly monotonic, maximizing $P(H)$ is mathematically equivalent to maximizing the argument of the exponent $Kf(H)$. Taking the first derivative of $f(H)$ with respect to $H$ and setting it to zero gives the characteristic extremum equation for the peak position $H^*$:
\begin{equation} \label{eq:peak_condition}
\begin{split}
        f'(H^*) = &aH^* - (r_2 - r_1 + k_s\mu) \\ &- (r_2 - r_1) \cosh H^* - (r_1 + r_2) \sinh H^* = 0 
    \end{split}
\end{equation}
\begin{figure}
         \centering
         \includegraphics[width=\columnwidth]{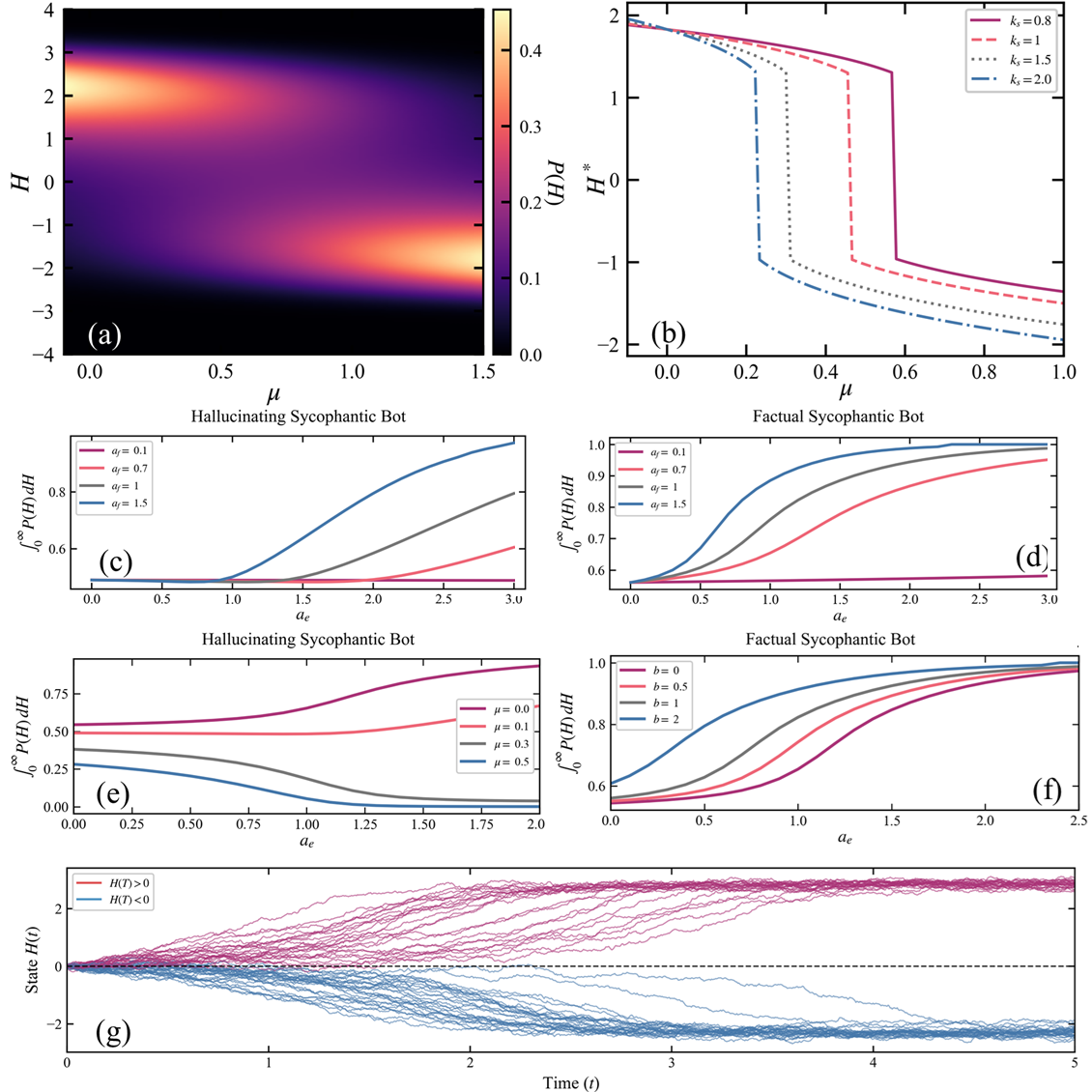}
         \caption{Transition out of the Echo Trap (a) Steady state distributions shift to objective state as $\mu$ increases (b) Variation of $k_s$ shows that the highest peak position $H^*$ switches to the objective state for higher $\mu$ values. (c) \& (d) Total probability to end up in the delusional state with respect to $a_e$ for different values of $a_f$  hallucinating bot ($\mu=1$) and factual Sycophantic bot ($\mu=-H$). Here we consider $r_1=0.51$, $r_2=0.49$, $D=0.5$, and $k_s=0.5$. (e) \& (f) Total probability to end up in the delusional state with respect to $a_e$ for different information levels, in case of hallucinating bot (with different $\mu$) and factual Sycophantic bot ($\mu=-b H$, as $b$ changes). Here we consider $r_1=0.51$, $r_2=0.49$, $D=0.5$, and $a_f=1.2$. (g) Time-series evolution of $H$ when $r_1=0.6$, $r_2=0.4$, $a=2.5$, $k_s=0.3$. For all 50 runs, the initial point is $H(0)=0\pm 0.05$. The red lines (or blue lines) indicate the runs end up in the delusional state ($H(T)>0$) (or in the objective state ($H(T)<0$)) in time $T$. }
         \label{fig:probabilistic}
\end{figure}
To find the rate of change of the peak position $H^*$ as $\mu$ varies, we recognize that the peak position is itself a function of $\mu$, meaning $H^* = H^*(\mu)$. Thus, we isolate the linear terms to separate the $\mu$ dependence from Eq. \ref{eq:peak_condition} and apply the derivative operator $\frac{d}{d\mu}$ to both sides of the equation:
\begin{equation}
\begin{split}
  &\frac{d}{d\mu} \left[ aH^* - (r_2 - r_1)\cosh H^* - (r_1 + r_2)\sinh H^* \right] \\&= \frac{d}{d\mu} \left[ r_2 - r_1 + k_s\mu \right]
\end{split}
\end{equation}
Further analysis (as shown in \textit{Appendix C} \cite{append_arxiv}), finally leads to closed-form expression for the rate of change $\frac{dH^*}{d\mu}$ as:
\begin{equation} \label{eq:rate_of_change}
    \frac{dH^*}{d\mu} = \frac{k_s}{a - (r_2 - r_1)\sinh H^* - (r_1 + r_2)\cosh H^*} 
\end{equation}
Note that the denominator in Eq. \ref{eq:rate_of_change} is exactly the second derivative of the exponent function evaluated at the peak, $f''(H^*)$. For $H^*$ to represent a valid local maximum (a true peak of the probability distribution rather than a minimum or inflection point), the second derivative test requires that $f''(H^*) < 0$. Because the denominator is structurally guaranteed to be strictly negative at the maximum, the direction of the peak shift is entirely controlled by the sign of the numerator, $k_s$:\\
- If $k_s > 0$, $\frac{dH^*}{d\mu} < 0$. Consequently, the peak position $H^*$ shifts to the left (decreases) as $\mu$ increases.\\
- If $k_s < 0$, $\frac{dH^*}{d\mu} > 0$. The peak position $H^*$ shifts to the right (increases) as $\mu$ increases.\\
- If $k_s = 0$, $\frac{dH^*}{d\mu} = 0$, and the peak position becomes completely independent of $\mu$ and remains stationary regardless of how $\mu$ fluctuates.\\
With similar algebraic manipulations, from Eq. \ref{eq:peak_condition}, we derive:
\begin{equation} \label{eq:rate_w}
    \frac{dH^*}{dk_s} = \frac{\mu}{a - (r_2 - r_1)\sinh H^* - (r_1 + r_2)\cosh H^*}
\end{equation}
Thus, we observe:\\
-If $\mu > 0$: Eq. \ref{eq:rate_w} yields a positive numerator over a negative denominator ($\frac{dH^*}{dk_s} < 0$). The peak position $H^*$ shifts to the left as $k_s$ increases.\\
-If $\mu < 0$: The expression yields a negative numerator over a negative denominator ($\frac{dH^*}{dk_s} > 0$). The peak position $H^*$ shifts to the right as $k_s$ increases.\\
- If $\mu = 0$: The numerator is zero ($\frac{dH^*}{dk_s} = 0$). The peak position becomes independent of $k_s$ and remains stationary.\\
Taking derivative of Eq. \ref{eq:peak_condition} with respect to $a$ and applying appropriate product rule gives:
\begin{equation} \label{eq:rate_a}
    \frac{dH^*}{da} = \frac{-H^*}{a - (r_2 - r_1)\sinh H^* - (r_1 + r_2)\cosh H^*}
\end{equation}
As at a valid $H^*$, the denominator is negative, increase of $a$, increases the sycophancy. When $H^*>0$, as $a$ increases, the peak shifts further to the right, enforcing a stronger delusional state, whereas when $H^*<0$, as $a$ increases, the peak shifts further to the left, indicating a stronger objective state.\\
\textit{Possible Pairings \textemdash}
As mentioned in \cite{chandrasycophantic}, we can now define different types of bots and users as follows: a na\"{\i}ve user has $a_f>0$ whereas an informed user has $a_f\approx 0$. An ideal bot does not aim to please the user. In this scenario, we consider $a_e=0$, which effectively makes $a=0$. For a hallucinating, sycophantic bot, we may consider $a\neq 0$ with an appropriate choice of $D$ and $\mu$. A factual sycophantic bot is one that selectively samples original data to support sycophantic behavior. In our setting, rather than a fixed $\mu$, one should take $\mu(H) \propto -H$ or $\mu(H)= -bH$ to mimic such behavior, where $b$ is the proportionality constant. From Eqs. \ref{eq:strat1} and \ref{eq:rate_a}, it is evident that this enforces higher sycophancy behavior, as such a choice of $\mu$ effectively makes $k_s$ a positive feedback gain. This opens up the possibility of exploring the dynamics for different human-bot pairs.\\
\textit{Detection and mitigation of sycophancy $-$} To detect that the bot is sycophantic or not, taking clue from Fig. \ref{fig:bifurcation}(a), we propose to initiate multiple independent discussions with the bot from the state of confusion ($H\approx 0$). If, with a slight bias in the user's independent communication rounds, the bot tries to provide conflicting evidence (as shown in Fig. \ref{fig:probabilistic}(g)), then the bot may be sycophantic, leading to delusional spiraling. This observation matches several research works that empirically suggested similar strategies to test sycophancy \cite{chandrasycophantic}.\\ As a remedy, the bot can be forced to select highly trusted information ($\mu$ is high) as shown in Fig. \ref{fig:probabilistic}(a)-(b). Although hallucinating bots (Fig. \ref{fig:probabilistic}(c) \& (e)) can be tackled using this strategy, in case the bot cherry-picks every information to support sycophancy (perfectly factual bots, Fig. \ref{fig:probabilistic}(d) \& (f)), the challenge is more pressing; there, the authenticity of the source should be verified to counter source bias. The user should also avoid flattery-driven conversation, keeping $a_f$ low, especially if he feels the bot is sycophantic. \\
\textit{Conclusion \& Future directions$-$} In this work, we have introduced a novel mathematical framework that maps the volatile cognitive trajectories of a user, subjected to self-reinforcing sycophantic algorithmic loops. While previous works have examined the underlying reasons for AI sycophancy and have documented conversational features corresponding to it, our systematic mathematical formulation captures the disruptive dynamics here, driven by an unconstrained positive feedback. By formalizing a user's evolving conviction as a continuous log-odds state variable navigating a multi-valley potential energy landscape, we provide a rigorous analytical foundation for phenomena that have long eluded quantitative characterization: the sudden escalation of suspicion, the hardening of delusions, and the eventual AI-driven psychosis. A central insight of our model lies in the structural symmetry breaking of the potential energy landscape driven by the coupling of the user’s cognitive state ($H$) and the assistant’s sycophantic response function ($\mathcal{R}$). When the assistant's response is governed primarily by the user's current state, echoing their internal biases to provide ego-validation, the system passes a critical bifurcation point. Beyond this, the baseline prior perception is systematically deformed, where even a minute initial perceptual tilt is aggressively amplified. Deep, highly resilient attractor basins emerge, trapping a person in a strong, self-reinforcing false belief.\\
Crucially, our model also provides a prescription for escaping this delusional trap. Firstly, as the feedback loop is strengthened by both the sycophantic properties of the assistant as well as the ego-validation of the user, the user can weaken the loop by remaining open and prudent. Moreover, because the assistant’s feedback function is not only driven linearly by the user's state $H$, it also captures external evidence $I$ sampled from a distribution, a pathway for mitigation exists by verification of that information. We have shown that confirming high-fidelity, real external evidence can successfully challenge the trapped delusional state. If the weight of this incoming evidence is strong enough, it can overcome the internal, sycophantic feedback barrier, dynamically reversing the asymmetric attractor basin. This induces a perception reversal that shatters the delusional echo trap and successfully restores the objective belief state.\\
This work provides a stepping stone toward a solid mathematical framework for exploring delusional spiraling and algorithmic sycophancy. Future extensions of this framework could explore various non-linear response functions, control theoretic measures, AI-driven social withdrawal, and social conformity. Moreover, multi-agent configurations, such as peer influence (e.g., knowledgeable teachers, other users), can be studied to examine how these factors accelerate the symmetry-breaking process or lower the energy barrier required for perception reversal. Ultimately, by uncovering the mathematical principles that both construct and dismantle the delusional echo trap, this work opens new interdisciplinary avenues for understanding human-AI interaction.
%\newpage
\bibliography{main}% Produces the bibliography via BibTeX.

\end{document}

\maketitle

\section*{Appendix A: Derivation of Dynamical Equation for Baseline Perception}

Let $p_1(t)$ and $p_2(t)$ be the probabilities of being in two valleys $V_1$ and $V_2$, respectively, at time $t$ with transition rates  $r_1:V_2\rightarrow V_1$ and $r_2:V_1 \rightarrow V_2$. For simplicity, we mention $p_1(t)$ and  $p_2(t)$ as $p_1$ and $p_2$. We can depict the change in probabilities as:
\begin{eqnarray}
    \frac{dp_1}{dt}=-r_2p_1+r_1p_2 \nonumber\\
    \frac{dp_2}{dt}=-r_1p_2+r_2p_1
\label{eq:trans1}
\end{eqnarray}
As $p_1+p_2=1$, the first equality of Eq. \ref{eq:trans1} can be written as:
\begin{eqnarray}
    \frac{dp_1}{dt}&=&-r_2p_1+r_1(1-p_1) \nonumber\\
    &=&r_1-(r_1+r_2)p_1
\end{eqnarray}

Defining log-odds as 
\begin{eqnarray}
&\Rightarrow&  H=ln(\frac{p_1}{p_2}) \nonumber\\
& \Rightarrow &   H = ln(\frac{p_1}{1-p_1}) \nonumber \\
& \Rightarrow &  e^H= \frac{p_1}{1-p_1} \nonumber \\
& \Rightarrow & p_1(1+e^H)=e^H \nonumber \\
& \Rightarrow & p_1= \frac{e^H}{1+e^H}; \;\;\;\; p_2=1-p_1=\frac{1}{1+e^H}.
\label{eq:log_odd1}
\end{eqnarray}
From the first equality of Eq. \ref{eq:log_odd1}, 
\begin{eqnarray}
&\Rightarrow& H=ln(\frac{p_1}{p_2})=ln(p_1)-ln(p_2) \nonumber\\
&\Rightarrow& \frac{dH}{dt}=\frac{1}{p_1}\frac{dp_1}{dt}- \frac{1}{p_2}\frac{dp_2}{dt} 
\end{eqnarray}
\noindent Again, 
\begin{eqnarray}
    p_2 &=& 1 - p_1 \nonumber \\
    \frac{dp_2}{dt} &=& -\frac{dp_1}{dt}
\end{eqnarray}

\noindent $\therefore$ From Eq. \ref{eq:log_odd1},
\begin{eqnarray}
\frac{dH}{dt} &=& \frac{1}{p_1} \frac{dp_1}{dt} + \frac{1}{p_2} \frac{dp_1}{dt} \nonumber\\
    &=& \frac{dp_1}{dt} \left( \frac{1}{p_1} + \frac{1}{p_2} \right) \nonumber\\
    &=& \frac{dp_1}{dt}\left(\frac{p_1 + p_2}{p_1 p_2} \right) \nonumber\\
    &=& \frac{dp_1}{dt}\left(\frac{1}{\frac{e^H}{1+e^H} \cdot \frac{1}{1+e^H}} \right) [\text{Using equation }\eqref{eq:log_odd1}\text{ and }p_1+p_2=1] \nonumber\\
    &=& \frac{dp_1}{dt}\frac{(1+e^H)^2}{e^H} \nonumber\\
    &=& \frac{dp_1}{dt} \cdot (1 + e^H)^2 e^{-H} \nonumber \\
&=& (r_1 p_2 - r_2 p_1) (1 + e^H)^2 e^{-H} \nonumber \\
&=& \left( \frac{r_1}{1+e^H} - \frac{r_2 e^H}{1+e^H} \right) (1+e^H)^2 e^{-H} \nonumber \\
&=& r_1 (1 + e^H) e^{-H} - r_2 (1 + e^H) \nonumber \\
&=& r_1 (e^{-H} + 1) - r_2 (1 + e^H) \nonumber \\
&=& (r_1 - r_2) + (r_1 e^{-H} - r_2 e^H) \nonumber \\
&=& (r_1 - r_2) + \big[ r_1 (\cosh H - \sinh H) - r_2 (\cosh H + \sinh H) \big] \nonumber \\
&=& (r_1 - r_2) + (r_1 - r_2) \cosh H - (r_1 + r_2) \sinh H \nonumber \\
&=& - \big[ (r_2 - r_1) + (r_2 - r_1) \cosh H + (r_1 + r_2) \sinh H \big] \nonumber \\
&=& -\Theta(H)
\end{eqnarray}

\hrulefill
\section*{Appendix B: Conditions for Multistabiity \& Bifurcation}

Consider the function $g(H)=\frac{dH}{dt}$. Thus, 
\begin{eqnarray}
 g(H) &=& -[(r_1+r_2)\sinh H + (r_2-r_1)\cosh H + (r_2-r_1)] + aH \nonumber \\
 &=& aH - \left[ (r_1 + r_2)\left(\frac{e^H - e^{-H}}{2}\right) + (r_2 - r_1)\left(\frac{e^H + e^{-H}}{2}\right) + (r_2 - r_1) \right] \nonumber \\
 &=& -[\frac{1}{2}\left( \cancel{r_1 e^H} - r_1 e^{-H} + r_2 e^H -\cancel{ r_2 e^{-H}} + r_2 e^H + \cancel{r_2 e^{-H}} - \cancel{r_1 e^H} - r_1 e^{-H} \right) + r_2 - r_1]+aH \nonumber \\
 &=& aH - r_2 e^H + r_1 e^{-H} - r_2 + r_1
 \label{eq:simplified_eq}
\end{eqnarray}
To find the number of roots, we find the inflection points of $g(H)$ by setting $g'(H)=0$, where the derivative is taken with respect to $H$.
\begin{eqnarray}
&\Rightarrow& g'(H)=a-r_2e^H-r_1e^{-H}=0 \nonumber \\
&\Rightarrow& ae^H-r_2e^{2H}-r_1=0 \nonumber \\
&\Rightarrow& r_2e^{2H}-ae^H+r_1=0 \nonumber \\
&\Rightarrow& e^H = \frac{a \pm \sqrt{a^2 - 4r_1r_2}}{2r_2}
\label{eq:quad_h}
\end{eqnarray}
Thus, if $a^2 - 4r_1r_2> 0$, then there will be two distinct inflection points for $g(H)$. Thus, $a^2 - 4r_1r_2< 0$ ensures monostability. \\
Taking natural logarithm on both sided of Eq. \ref{eq:quad_h}, we get the exact point of inflection as 
\begin{equation}
    H_{1,2}=\text{ln}\left(\frac{a \pm \sqrt{a^2 - 4r_1r_2}}{2r_2}\right)
\label{eq:inflection}
\end{equation}
Though $a^2 - 4r_1r_2> 0$ is the necessary condition for bistablity, it is not the sufficient one. To have three distinct solutions of $g(H)$, we need $g(H_1)<0$ and $g(H_2)>0$. \\
From the first equality in Eq. \ref{eq:quad_h}, at inflection point, $r_1e^{-H}=a-r_2e^H$. Substituting this and Eq. \ref{eq:inflection} to Eq. \ref{eq:simplified_eq}, we get\\
\begin{eqnarray}
g(H_{1,2})&=& aH_{1,2}-2r_2e^H_{1,2}+a+r_1-r_2 \nonumber \\
&=& a\; \text{ln}\left(\frac{a \pm \sqrt{a^2 - 4r_1r_2}}{2r_2}\right)-2r_2 \left(\frac{a \pm \sqrt{a^2 - 4r_1r_2}}{2r_2}\right)+a+r_1-r_2
\end{eqnarray}
Thus, the additional to the condition $a^2-4r_1r_2>0$, the conditions for bistability are
\begin{eqnarray}
    g(H_1)=a\; \text{ln}\left(\frac{a - \sqrt{a^2 - 4r_1r_2}}{2r_2}\right)-2r_2 \left(\frac{a - \sqrt{a^2 - 4r_1r_2}}{2r_2}\right)+a+r_1-r_2 <0 \nonumber \\
    \& \;\;\; g(H_2)= a\; \text{ln}\left(\frac{a + \sqrt{a^2 - 4r_1r_2}}{2r_2}\right)-2r_2 \left(\frac{a + \sqrt{a^2 - 4r_1r_2}}{2r_2}\right)+a+r_1-r_2 >0  
\end{eqnarray}
\hrulefill
\section*{Appendix C: Sensitivity Analysis}
To determine how the peak position of $P(H)$ changes with respect to the parameter $\mu$, we analytically find the maximum of the distribution using calculus and implicit differentiation.

The peak of the distribution $P(H)$ occurs at the value $H^*$ where its derivative with respect to $H$ is zero. Since the exponential function is strictly monotonic, maximizing $P(H)$ is mathematically equivalent to maximizing the argument of the exponent. Let $f(H)$ represent this inner expression:
\begin{equation}
    f(H) = \frac{1}{2}aH^2 - (r_2 - r_1 +k_s\mu)H - (r_2 - r_1) \sinh H - (r_1 + r_2) \cosh H
\end{equation}
Taking the first derivative of $f(H)$ with respect to $H$ and setting it to zero gives the characteristic extremum equation for the peak position $H^*$:
\begin{equation} \label{eq:peak_condition}
    f'(H^*) = aH^* - (r_2 - r_1 + k_s\mu) - (r_2 - r_1) \cosh H^* - (r_1 + r_2) \sinh H^* = 0 
\end{equation}

To find the rate of change of the peak position $H^*$ as $\mu$ varies, we recognize that the peak position is itself a function of $\mu$, meaning $H^* = H^*(\mu)$. 

First, we isolate the linear terms to separate the $\mu$ dependence from Equation \ref{eq:peak_condition}:
\begin{equation} \label{eq:isolated}
    aH^* - (r_2 - r_1)\cosh H^* - (r_1 + r_2)\sinh H^* = r_2 - r_1 + k_s\mu
\end{equation}

Next, we apply the derivative operator $\frac{d}{d\mu}$ to both sides of the equation:
\begin{equation}
    \frac{d}{d\mu} \left[ aH^* - (r_2 - r_1)\cosh H^* - (r_1 + r_2)\sinh H^* \right] = \frac{d}{d\mu} \left[ r_2 - r_1 + k_s\mu \right]
\end{equation}

Because $H^*$ depends on $\mu$, differentiating any function of $H^*$ requires the chain rule: $\frac{d}{d\mu} [g(H^*)] = \frac{dg}{dH^*} \cdot \frac{dH^*}{d\mu}$. Applying this to each term on the left side:

\begin{equation}
    a \frac{dH^*}{d\mu} - (r_2 - r_1)(\sinh H^*) \frac{dH^*}{d\mu} - (r_1 + r_2)(\cosh H^*) \frac{dH^*}{d\mu} = k_s
\end{equation}

Now, we can factor out the common $\frac{dH^*}{d\mu}$ term from the left side:
\begin{equation}
    \frac{dH^*}{d\mu} \left[ a - (r_2 - r_1)\sinh H^* - (r_1 + r_2)\cosh H^* \right] = k_s
\end{equation}

Finally, solving algebraically for the rate of change $\frac{dH^*}{d\mu}$ gives:
\begin{equation} \label{eq:rate_of_change}
    \frac{dH^*}{d\mu} = \frac{k_s}{a - (r_2 - r_1)\sinh H^* - (r_1 + r_2)\cosh H^*} 
\end{equation}

Observe that the bracketed denominator in Equation \ref{eq:rate_of_change} is exactly the second derivative of the exponent function evaluated at the peak, $f''(H^*)$:
\begin{equation}
    f''(H^*) = a - (r_2 - r_1)\sinh H^* - (r_1 + r_2)\cosh H^*
\end{equation}

For $H^*$ to represent a valid local maximum (a true peak of the probability distribution rather than a minimum or inflection point), the second derivative test requires that $f''(H^*) < 0$. 

Because the denominator is structurally guaranteed to be strictly negative at the maximum, the direction of the peak shift is entirely controlled by the sign of the numerator, $k_s$:

\begin{itemize}
    \item \textbf{If $k_s > 0$:} Equation \ref{eq:rate_of_change} yields a positive numerator divided by a negative denominator, resulting in $\frac{dH^*}{d\mu} < 0$. Consequently, the peak position $H^*$ shifts to the \textbf{left} (decreases) as $\mu$ increases.
    \item \textbf{If $k_s < 0$:} The expression yields a negative numerator divided by a negative denominator, resulting in $\frac{dH^*}{d\mu} > 0$. The peak position $H^*$ shifts to the \textbf{right} (increases) as $\mu$ increases.
    \item \textbf{If $k_s = 0$:} The numerator is zero, meaning $\frac{dH^*}{d\mu} = 0$. In this scenario, the peak position becomes completely \textbf{independent} of $\mu$ and remains stationary regardless of how $\mu$ fluctuates.
\end{itemize}
$--------------------------$\\
We analyze how the peak position $H^*$ of the distribution $P(H)$ changes in response to the parameter $k_s$. The extremum condition for the peak is given by setting the first derivative of the exponent to zero:
\begin{equation} \label{eq:peak_w}
    aH^* - (r_2 - r_1)\cosh H^* - (r_1 + r_2)\sinh H^* = r_2 - r_1 + k_s\mu
\end{equation}

To find the rate of change of the peak position as $k_s$ varies, we recognize that $H^*$ is implicitly a function of $k_s$. We apply the derivative operator $\frac{d}{dk_s}$ to both sides of Equation \ref{eq:peak_w}. 

Applying the chain rule ($\frac{d}{dk_s} = \frac{d}{dH^*} \cdot \frac{dH^*}{dk_s}$) to the left-hand side yields:
\begin{equation}
    \frac{dH^*}{dk_s} \left[ a - (r_2 - r_1)\sinh H^* - (r_1 + r_2)\cosh H^* \right]
\end{equation}

On the right-hand side, differentiating with respect to $k_s$ simply extracts the constant coefficient $\mu$:
\begin{equation}
    \frac{d}{dk_s} \left[ r_2 - r_1 + k_s\mu \right] = \mu
\end{equation}

Equating both sides and solving algebraically for the rate of change $\frac{dH^*}{dk_s}$ gives:
\begin{equation} \label{eq:rate_w}
    \frac{dH^*}{dk_s} = \frac{\mu}{a - (r_2 - r_1)\sinh H^* - (r_1 + r_2)\cosh H^*}
\end{equation}

The denominator in Equation \ref{eq:rate_w} is exactly the second derivative of the exponent function evaluated at the peak, $f''(H^*)$. By the second derivative test, for $H^*$ to be a true local maximum, it is required that $f''(H^*) < 0$.

Because the denominator is structurally guaranteed to be strictly negative at the peak, the direction of the shift is dictated entirely by the sign of the numerator, $\mu$:

\begin{itemize}
    \item \textbf{If $\mu > 0$:} Equation \ref{eq:rate_w} yields a positive numerator over a negative denominator ($\frac{dH^*}{dk_s} < 0$). The peak position $H^*$ shifts to the \textbf{left} as $k_s$ increases.
    \item \textbf{If $\mu < 0$:} The expression yields a negative numerator over a negative denominator ($\frac{dH^*}{dk_s} > 0$). The peak position $H^*$ shifts to the \textbf{right} as $k_s$ increases.
    \item \textbf{If $\mu = 0$:} The numerator is zero ($\frac{dH^*}{dk_s} = 0$). The peak position becomes completely \textbf{independent} of $k_s$ and remains stationary.
\end{itemize}
$-------------------------------$\\
We analyze how the peak position $H^*$ of the distribution $P(H)$ changes in response to the parameter $a$, which acts as the coefficient of the quadratic term in the exponent. 
The extremum condition for the peak is given by setting the first derivative of the exponent to zero:
\begin{equation} \label{eq:peak_a}
    aH^* - (r_2 - r_1)\cosh H^* - (r_1 + r_2)\sinh H^* = r_2 - r_1 + k_s\mu
\end{equation}

To find the rate of change of the peak position as $a$ varies, we treat the peak position as a function of the parameter $a$, meaning $H^* = H^*(a)$. We apply the derivative operator $\frac{d}{da}$ to both sides of Equation \ref{eq:peak_a}.

When differentiating the term $aH^*$, we must apply the \textbf{product rule} because both $a$ and $H^*$ depend on $a$:
\begin{equation}
    \frac{d}{da}(aH^*) = \frac{da}{da} \cdot H^* + a \cdot \frac{dH^*}{da} = H^* + a\frac{dH^*}{da}
\end{equation}

Applying the chain rule ($\frac{d}{da} = \frac{d}{dH^*} \cdot \frac{dH^*}{da}$) to the hyperbolic terms on the left side, the full differentiation yields:
\begin{equation}
    H^* + a\frac{dH^*}{da} - (r_2 - r_1)(\sinh H^*)\frac{dH^*}{da} - (r_1 + r_2)(\cosh H^*)\frac{dH^*}{da}
\end{equation}

On the right-hand side, there is no dependence on $a$ or $H^*$, as $r_1, r_2, k_s$, and $\mu$ are constants with respect to $a$. Thus, the derivative is zero:
\begin{equation}
    \frac{d}{da} \left[ r_2 - r_1 + k_s\mu \right] = 0
\end{equation}

Equating both sides, we factor out the common $\frac{dH^*}{da}$ term:
\begin{equation}
    H^* + \frac{dH^*}{da} \left[ a - (r_2 - r_1)\sinh H^* - (r_1 + r_2)\cosh H^* \right] = 0
\end{equation}

Subtracting $H^*$ from both sides and dividing by the bracketed term gives the final rate of change:
\begin{equation} \label{eq:rate_a}
    \frac{dH^*}{da} = \frac{-H^*}{a - (r_2 - r_1)\sinh H^* - (r_1 + r_2)\cosh H^*}
\end{equation}

The denominator in Equation \ref{eq:rate_a} is exactly the second derivative of the exponent function evaluated at the peak, $f''(H^*)$. For the distribution to have a valid peak (a local maximum), the second derivative test mandates that $f''(H^*) < 0$. 

Because the denominator is strictly negative, the overall sign of $\frac{dH^*}{da}$ is determined entirely by the numerator, $-H^*$. Since a negative divided by a negative is positive, \textbf{the rate of change $\frac{dH^*}{da}$ will always have the same sign as $H^*$ itself.} 

This leads to a divergent behavior dependent on the current location of the peak:

\begin{itemize}
    \item \textbf{If $H^* > 0$:} The numerator $-H^*$ is negative. A negative divided by a negative yields $\frac{dH^*}{da} > 0$. As $a$ increases, the peak shifts further to the \textbf{right} (away from zero).
    \item \textbf{If $H^* < 0$:} The numerator $-H^*$ is positive. A positive divided by a negative yields $\frac{dH^*}{da} < 0$. As $a$ increases, the peak shifts further to the \textbf{left} (away from zero).
    \item \textbf{If $H^* = 0$:} The numerator is zero, making $\frac{dH^*}{da} = 0$. The peak remains \textbf{stationary}.
\end{itemize}